\documentclass[10pt,twocolumn,letterpaper]{article}

\usepackage[final,applications]{wacv}      

\usepackage{graphicx}
\usepackage{amsmath}
\usepackage{amssymb}
\usepackage{booktabs}

\usepackage[pagebackref,breaklinks,colorlinks]{hyperref}

\usepackage[capitalize]{cleveref}
\crefname{section}{Sec.}{Secs.}
\Crefname{section}{Section}{Sections}
\Crefname{table}{Table}{Tables}
\crefname{table}{Tab.}{Tabs.}

\def\wacvPaperID{1344} 
\def\confName{WACV}
\def\confYear{2025}

\def\ours{ConDiSR}
\begin{document}

\title{ConDiSR: Contrastive Disentanglement and Style Regularization for Single Domain Generalization}

\author{Aleksandr Matsun\\
MBZUAI\\
United Arab Emirates\\
{\tt\small aleksandr.matsun@mbzuai.ac.ae}
\and
Numan Saeed\\
MBZUAI\\
United Arab Emirates\\
{\tt\small numan.saeed@mbzuai.ac.ae}
\and
Fadillah Adamsyah Maani\\
MBZUAI\\
United Arab Emirates\\
{\tt\small fadillah.maani@mbzuai.ac.ae}
\and
Mohammad Yaqub\\
MBZUAI\\
United Arab Emirates\\
{\tt\small mohammad.yaqub@mbzuai.ac.ae}
}
\maketitle

\begin{abstract}
   Medical data often exhibits distribution shifts, leading to performance degradation of deep learning models trained using standard supervised learning pipelines. Domain Generalization (DG) addresses this challenge, with Single-Domain Generalization (SDG) being notably relevant due to the privacy and logistical constraints often inherent in medical data. Existing disentanglement-based SDG methods heavily rely on structural information from segmentation masks, but classification labels do not offer similarly dense information.
   This work introduces a novel SDG method for medical image classification, utilizing channel-wise contrastive disentanglement. The method is further refined with reconstruction-based style regularization to ensure distinct style and structural feature representations are extracted. We evaluate our method on the complex tasks of multicenter histopathology image classification and Diabetic Retinopathy (DR) grading in fundus images, benchmarking it against state-of-the-art (SOTA) SDG baselines. Our results demonstrate that our method consistently outperforms the SOTA independently on the choice of the source domain while exhibiting greater performance stability. This study underscores the importance and challenges of exploring SDG frameworks for classification tasks. The code is publicly available at \url{https://github.com/BioMedIA-MBZUAI/ConDiSR}
\end{abstract}

\section{Introduction}
\label{sec:intro}
In recent years, medical imaging analysis has experienced significant advancements powered by deep learning \cite{varoquaux2022machine}. 
These models are typically trained under the assumption of independent and identically distributed (i.i.d.) data samples. However, this assumption can significantly hamper performance during inference in real-world scenarios due to distribution shifts. This issue is particularly significant in medical imaging data due to differences between medical centers, variations in equipment, and the inherent complexity and variability of biological structures. Using large and diverse datasets during training to address these distribution shifts is essential. However, obtaining such datasets is challenging due to privacy concerns \cite{DG_Survey}.

To address these constraints, the training process can be arranged in a Domain Adaptation (DA) \cite{kouw2019introduction} or Domain Generalization (DG) \cite{domainbed} framework. DA focuses on adapting a model trained on one or more source domains to perform well on a target domain, which may have a different distribution. On the other hand, DG aims to develop models that generalize well across unseen domains without having access to data from those (target) domains during training. These approaches assume that data can be divided into distinct domains, within which samples follow the i.i.d. assumption. Applying these techniques makes it possible to develop machine-learning models that exhibit improved performance and generalizability, even with the limited availability of heterogeneous data. 

In the setting of Unsupervised DA (UDA) \cite{wilson2020survey,zhang2020collaborative}, training is performed using data from the source domain(s) as well as unlabeled data from the target domain. Conversely, the Multi-Source DG (MSDG) \cite{DG_Survey,li2020domain} setting involves training using data from multiple source domains and evaluating on unseen target domain(s). However, obtaining data from multiple sources and target domains is often infeasible, especially in medical imaging. 

On the other hand, Single Domain Generalization (SDG)  \cite{wilson2020survey,qiao2020learning,hu2023devil,ouyang2022causality,su2023rethinking} uses data from a single source domain during training, better mimicking real-world scenarios. Compared to UDA and MSDG, SDG presents a more challenging task, as using data from one source increases the likelihood of model overfitting and reduces its ability to generalize to other domains. Methods that can help learn a generalizable model using only one source domain include broadening the input space via image augmentations \cite{ouyang2022causality,su2023rethinking} or performing feature-level augmentations that avoid semantic perturbations of the content \cite{zhou2021domain,li2022uncertainty,zhang2024domain}. Recently, \cite{hu2023devil} proposed an SDG method that achieves domain-independent feature representation extraction by disentangling style and structure-related components based on the assumption that semantic information is primarily encoded in an image's high-frequency components. In contrast, style/texture-related information resides in its low-frequency components \cite{Wang_2022}. 

While \cite{hu2023devil} shows SOTA results in the segmentation task, it performs poorly when adapted to classification. This is because the segmentation model training pipeline involves dense ground truth labels that provide significant structural information, aiding the model's disentanglement module. In contrast, a standard classification model is only given one numeric value per image during training - the class label, which provides significantly less structural information. This issue is especially noticeable in classifying histopathology images as well as fundus images. For the former it is caused by significant style variations due to different staining procedures \cite{xu2022improved} while for the latter by high impact of scanning equipment settings.

In this work, we propose a novel method, \ours{}, that combines \textbf{Con}trastive \textbf{Di}sentanglement and \textbf{S}tyle \textbf{R}egularization for robust medical image classification in the SDG setting. Our key contributions are as follows: 

\begin{itemize}
    \item We develop a new SDG method that utilizes reconstruction-based style regularization for improved structure/style disentanglement.
    \item We develop a criterion suitable for the classification task that integrates reconstruction loss, contrastive loss, and cross-entropy loss functions.
    \item We propose a solution to the medical images classification task that improves over the previous SOTA methods in SDG on two DG datasets representing different medical imaging modalities.
\end{itemize}

\section{Methodology}

\begin{figure*}[t]
  \centering
   \includegraphics[width=0.8\linewidth]{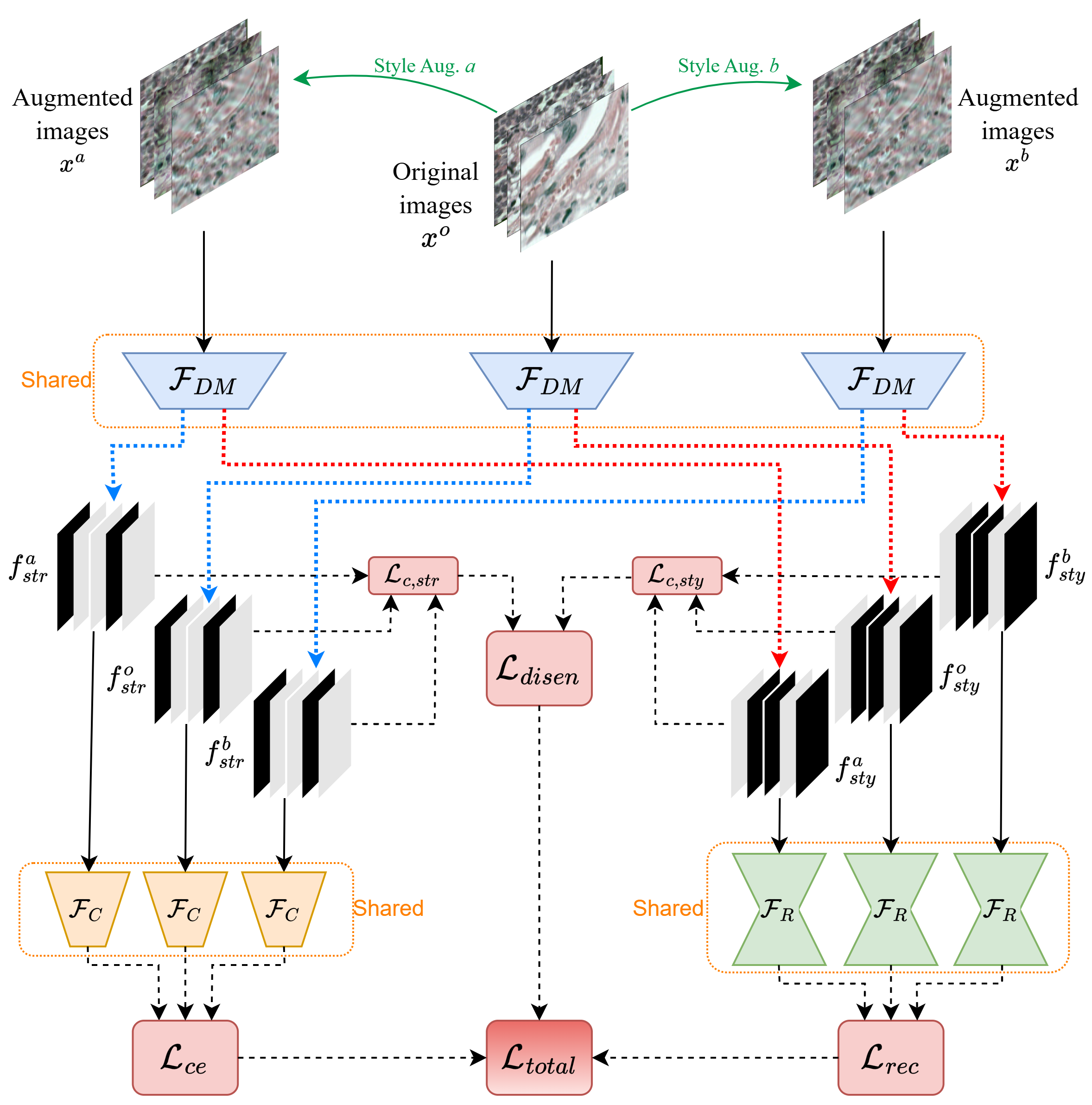}

   \caption{Overview of the proposed method. The feature disentanglement module ($\mathcal{F}_{DM}$) processes the original and augmented images to produce structure-related features ($f{str}$) and style-related features ($f_{sty}$). The structure-related features ($f_{str}$) are passed further to the classification network ($\mathcal{F}_C$) for computing the classification loss, while the style-related ones ($f_{sty}$) go through the reconstruction network ($\mathcal{F}_R$) for further computation of the reconstruction loss. Additionally a contrastive loss is applied to minimize the distance between similar structure-related components and maximize the distance between different style-related components.}
   \label{Figure:model}
\end{figure*}

\begin{figure}[t]
  \centering
   \includegraphics[width=0.7\linewidth]{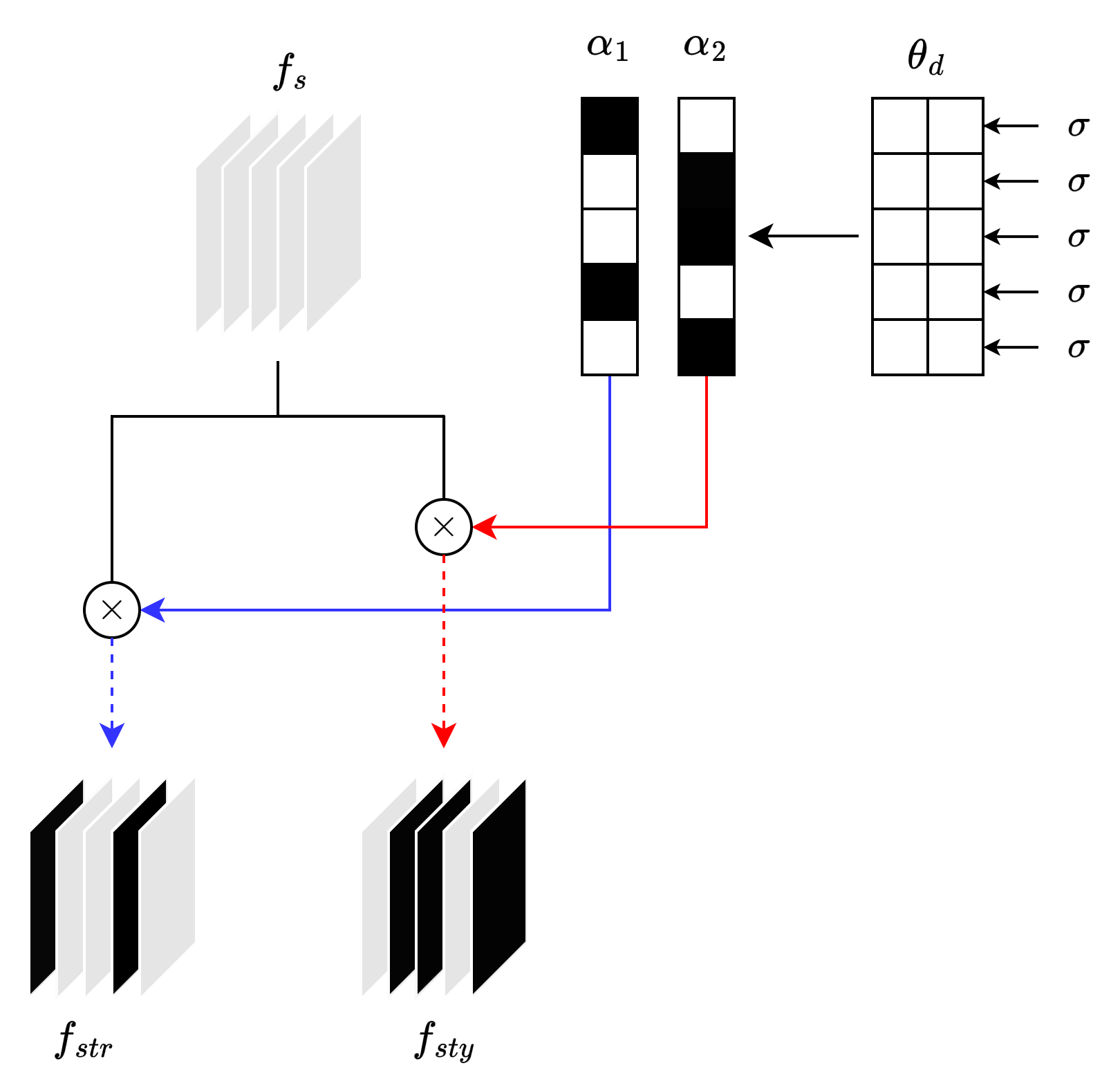}

   \caption{Overview of the feature disentanglement module of the network. Here \(\sigma\) represents the softmax operator, applied across the channel-related dimension of the parameter \(\theta_{d}\)}
   \label{Figure:model_fd}
\end{figure}

\subsection{Preliminaries}

\noindent \textbf{Problem Definition:} We first define $x \in \mathcal{X}$ as a medical image and $y \in \mathcal{Y}$ as its corresponding clinical label. Suppose that we have medical data from multiple domains or institutions, denoted as $\mathcal{D}=\{\mathcal{D}^n\}_{n=1}^{N_\mathcal{D}}$, where each institution $\mathcal{D}^n=\{x^n_i, y^n_i\}_{i=1}^{N_{\mathcal{D}^n}}$ typically represents different distributions due to various factors such as different machines and demographics. Our main aim is to learn a function that can accurately map $\mathcal{X} \to \mathcal{Y}$ using the available data while being robust to distribution shifts.

\noindent \textbf{Domain Generalization Setting:} In real-world practice, medical institutions are often reluctant to share their data due to privacy concerns. By assuming that we have access to data from only some institutions, we can decompose $\mathcal{D}$ into two sets:
\begin{equation}
    \mathcal{D} = \mathcal{S} \cup \mathcal{T}
\end{equation}
where $\mathcal{S}$ represents data from source institutions $\mathcal{S}=\{\mathcal{S}^n\}_{n=1}^{N_\mathcal{S}}$ and $\mathcal{T}$ represents data from target institutions $\mathcal{T}=\{\mathcal{T}^n\}_{n=1}^{N_\mathcal{T}}$. Consequently, we must rely on the source data to train a model that performs well on data from both seen and unseen distributions. Domain generalization aims to address this challenge.

\noindent \textbf{Single Domain Generalization (SDG):} This setting assumes that there is only a single domain available for training, i.e. $N_{\mathcal{S}}=1$. Developing a robust model in SDG is challenging due to the limited diversity of a single training data distribution.

\subsection{Proposed Method}

In order to overcome the challenges that the classification task in SDG setting poses, we propose \textit{\textbf{Con}trastive \textbf{Di}sentanglement and \textbf{S}tyle \textbf{R}egularization (ConDiSR)}, a reconstruction-based style regularization in conjunction with the technique of channel-wise contrastive disentanglement. Our proposed method is illustrated in Figure \ref{Figure:model}. Our model takes a medical image $x \in \mathbb{R}^{3 \times W \times H}$ as input and produces two outputs: \textbf{1)} $\hat{y}$, a prediction of its true clinical label $y \in \{0,1,\dots,C-1\}$, and \textbf{2)} a reconstructed image at a lower scale $\tilde{x} \in \mathbb{R}^{3 \times h \times w}$, used for reconstruction-based style regularization. Mathematically, the model can be represented as
\begin{equation}
    (\hat{y}, \tilde{x}) = \mathcal{F}(x)
\end{equation}

\subsubsection{Model Design}

We design our model $\mathcal{F}$ to mitigate severe performance degradation caused by such distribution shifts. The model $\mathcal{F}$ disentangles the features into style-rich features ($f_{sty}$) and structure-rich features ($f_{str}$). The style-rich features ($f_{sty}$) are used to reconstruct the input image, while the structure-rich features ($f_{str}$) are used for the classification task. This approach reduces the model's reliance on image style that varies across different institutions or domains, thereby improving robustness to distribution shifts. The model $\mathcal{F}$ comprises a \textit{feature disentanglement module} ($\mathcal{F_{DM}}$), a \textit{classification network} ($\mathcal{F}_C$), and a \textit{reconstruction network} ($\mathcal{F}_R$).

\noindent \textbf{Feature Disentanglement Module:} We design $\mathcal{F}_{DM}$ to effectively provide us with structure-rich features $f_{str}$ and style-rich features $f_{sty}$ from an input image $x$. The image $x$ is passed to the stem layer of the backbone convolutional network ($g=Conv2D(k=7,s=2,p=3)$) with $C_s$ output channels, to extract the shallow-level feature map, making the output more compact and enriched with distinct features. Let the output of the stem layer be denoted as $f_s \in \mathbb{R}^{C_s \times H/2 \times W/2}$. The channels of the feature map $f_s$ are believed to carry distinct style- and structure-related features \cite{Xie_2021}, so we perform the disentanglement by splitting these shallow level channels into structure- and style-related ones, thus comprising $f_{str}$ and $f_{sty}$. We implement this by introducing a learnable parameter $\theta_d \in \mathbb{R}^{2\times C_s}$ that represents the model's knowledge of which channels go to $f_{str}$ and which to $f_{sty}$ as shown on Figure \ref{Figure:model_fd}. Each row of $\theta_d$ corresponds to one channel of \(f_d\). In order to perform the disentanglement we separately apply softmax to each row of \(\theta_d\) which gives us two weight values per channel of \(f_d\). We use low temperature parameter in softmax to make sure that one of these weights closely approaches 0 while the other one - 1. By multiplying each of the channels with these weights, we facilitate the disentanglement of style and structure features. This approach allows the model to emphasize the contribution of each channel in providing style or structure information. Mathematically, the operations performed in $(f_{sty}, f_{str})=\mathcal{F}_{DM}(x)$ can be written as follows:
\begin{align}
    f_s &= g(x), &f_s &\in \mathbb{R}^{C_s \times H/2 \times W/2} \\
    \alpha_1, \alpha_2 &= \sigma_c(\theta_d / \tau), &\alpha &\in [0,1]^{C_s} \\
    f_{sty} &= \alpha_1 \ast f_s, &f_{sty} &\in \mathbb{R}^{C_s \times H/2 \times W/2} \\
    f_{str} &= \alpha_2 \ast f_s, &f_{str} &\in \mathbb{R}^{C_s \times H/2 \times W/2}
\end{align}
where $\sigma_c$ is the channel-wise softmax and $\tau \in \mathbb{R}$ is a temperature constant. By default, we set $C_s$ to $64$.

\noindent \textbf{Classification Network:} We harness the information embedded in structure-rich features $f_{str}$ to perform image classification, i.e. $\hat{y}=\mathcal{F}_C(f_{str})$. These features encapsulate essential structural information that is robust to distribution shifts and also potentially contain less but sufficient style information to aid in classification. By leveraging these comprehensive features, our classifier $\mathcal{F}_C$ is capable of effectively generalizing across different domains, thus maintaining high accuracy.

\noindent \textbf{Reconstruction Network:} In order to enhance performance of the disentanglement module and make sure that $f_{sty}$ indeed carries style-related features we impose an additional constraint on $f_{sty}$: it must carry enough information to reconstruct the lower-resolution version of the input image. We denote the reconstructed image as \( 
\tilde{x} = \mathcal{F}_{R}(f_{sty}) \). It is crucial that downsampled version of the $x$ is utilized as the reconstruction target in order to eliminate possible high-frequency (structural) features from it and thus avoid the possibility of $f_{sty}$ carrying structural information that could be necessary for the reconstruction of full-scale image. The style augmentations are also applied to the images that are used as the reconstruction targets, since we want $f_{sty}$ to carry the information that depends on the style-altering modifications.

\subsubsection{Training}

\noindent \textbf{Contrastive Disentanglement:} We utilize the concept of contrastive learning to train $\mathcal{F}_{DM}$ to effectively disentangle $f_{sty}$ and $f_{str}$ from $f_s$. For every original image $x$, we generate two augmented images using Bezier curve transformation from SLAug \cite{su2022rethinking} and low-frequency component replacement \cite{yang2020fda}. These two augmentation techniques alter the image style while preserving the original image structure. Let $x^o$ denote the original image, and ($x^a$, $x^b$) denote the augmented images. We then pass these images to the feature disentanglement module $\mathcal{F}_{DM}(\cdot)$ to produce their corresponding style-rich and structure-rich features, i.e. ($f_{sty}^o,f_{str}^o$), ($f_{sty}^a,f_{str}^a$), and ($f_{sty}^b,f_{str}^b$). The contrastive loss components are computed following the principle that, after disentanglement, \( f_{str}^{o}, f_{str}^{a} \) and \( f_{str}^{b} \) have to be the same as they represent the same structural pattern, while \( f_{sty}^{o}, f_{sty}^{a} \) and \( f_{sty}^{b} \) should be different due to the style alterations caused by augmentations. The contrastive disentanglement loss is then expressed as follows:
\begin{align}
\mathcal{L}_{c, str} &= \frac{1}{2} \cdot \sum_{\substack{ i,j \in \{o,a,b\} \\ i \neq j}} \left| p(f_{str}^i) - p(f_{str}^j) \right| \\
\mathcal{L}_{c, sty} &= - \frac{1}{2} \cdot \sum_{\substack{ i,j \in \{o,a,b\} \\ i \neq j}} \left| p(f_{sty}^i) - p(f_{sty}^j) \right| \\
\mathcal{L}_{disen} &= \mathcal{L}_{c, str} + \mathcal{L}_{c, sty}
\end{align}
where the $p: \mathbb{R}^{C_s \times H/2 \times W/2} \to \mathbb{R}^{C_c}$ is a learnable operator, performing projection of the feature maps into a latent space of smaller dimension \( C_c \). The projector is a small network, that consists of a convolution followed by a fully-connected layer. By default $C_c$ is set to 1024.

\noindent \textbf{Style Regularization:} The next step involves computation of the style regularization loss components, which have been specifically designed to impose additional constraints on the style-dependent features (\( f_{sty}^o, f_{sty}^a, f_{sty}^b \)) and improve the overall disentanglement capability of the architecture. To serve as the ground truth for the reconstruction, we utilize a downscaled version of the original images, which typically preserve style-related information and lose their inherent structural features due to resizing. We apply a reconstruction loss to compare the reconstructed images ($\tilde{x}^o,\tilde{x}^a,\tilde{x}^b$) with the downscaled images:
\begin{equation}
\mathcal{L}_{rec} = \sum_{i \in \{ o,a,b \}} \| \tilde{x}^i - rs(x^i) \|^2 ,
\end{equation}
where \(rs\) is the resizing operator.

\noindent \textbf{Final Objection Function:} Ultimately, the entire model $\mathcal{F}$ is trained by utilizing the total loss:
\begin{equation}
    \mathcal{L}_{total} = \mathcal{L}_{ce} + \mathcal{L}_{disen} + \mathcal{L}_{rec}
\end{equation}
where $\mathcal{L}_{ce}$ denotes the cross-entropy loss for supervised classification training.

\section{Experimental Setup}
\begin{figure*}[t]
\centering
\begin{tabular}{ccccc}
    \includegraphics[width=0.17\textwidth]{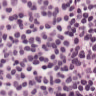} & \includegraphics[width=0.17\textwidth]{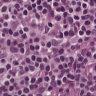} & \includegraphics[width=0.17\textwidth]{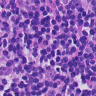} &
    \includegraphics[width=0.17\textwidth]{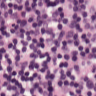} & \includegraphics[width=0.17\textwidth]{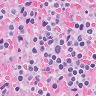} \\
    Center 0 & Center 1 & Center 2 & Center 3 & Center 4 \\
\end{tabular}
\caption{Sample images from the domains of the five centers of Camelyon17-WILDS.}
\label{fig:data}
\end{figure*}
\begin{figure*}[t]
\centering
\begin{tabular}{cccc}
    \includegraphics[width=0.17\textwidth]{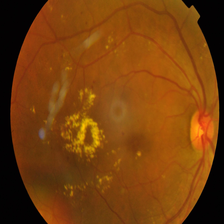} 
    & \includegraphics[width=0.17\textwidth]{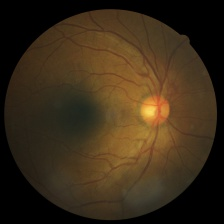} 
    & \includegraphics[width=0.17\textwidth]{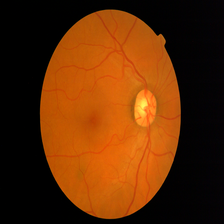} & \includegraphics[width=0.17\textwidth]{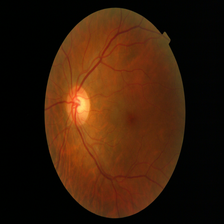} \\
    Center 0 & Center 1 & Center 2 & Center 3  \\
\end{tabular}
\caption{Sample images from the DR datasets.}
\label{fig:data_dr}
\end{figure*}
\textbf{Datasets.} The main experiments are conducted using the binary classification dataset Camelyon17-WILDS, which is a patchified version of the original Camelyon17 \cite{bandi2018detection} breast cancer whole-slide images dataset. The data was collected from five medical centers each representing a separate domain. All input images are sized \(96 \times 96 \), and the labels indicate whether the central \( 32 \times 32 \) area contains any tumor tissue. Additionally a composition of DR grading datasets Aptos \cite{aptos_ds}, EyePACS \cite{eyepacs_ds}, Messidor and Messidor2 \cite{messidor_ds} is used for testing the models' generalizability.

As depicted in Figures \ref{fig:data} and \ref{fig:data_dr}, sample images from the dataset showcase significant variations in texture throughout the different domains, making it a challenging task to develop a generalizable model.

\textbf{Evaluation Methods.} We use the DomainBed \cite{domainbed} evaluation protocol with training-domain validation model selection strategy. In order to properly estimate the generalization ability of the models in the SDG setting, we iteratively repeat training with each of the centers being set as the source domain while the union of the remaining four serve as the target domain.

\textbf{Networks architecture.} As the main backbone all the models utilize ResNet18 \cite{he2016deep} pre-trained on ImageNet \cite{deng2009imagenet}. Specifically the shallow feature map extraction operator \( g \) of the feature disentanglement module \( F_{DM} \) is implemented with the stem layer of ResNet18 while the classification network \( F_C \) is composed of its remaining layers. The reconstruction network \( F_R \) is implemented with a light-weight encoder-decoder model that involves 3 downsampling blocks and variable number of upsampling blocks depending on the reconstruction resolution.

\textbf{Implementation Details.} The experiments were run on 24GB Quadro RTX 6000 GPU. Each experiment was run for 50 epochs with the batch size of 256 using Adam optimizer \cite{kingma2014adam} with the learning rate of $1e-3$. The temperature parameter \( \tau \) was fixed at 0.1 similar to \cite{hu2023devil}.

We apply the same data augmentation strategy that we use in our method to all the other methods for a fair comparison. By doing so we eliminate the possibility of data augmentations being the deciding factor in the model's performance and equalize the total number images each model is fed during training.

\textbf{Ablation studies.} We conduct a study to investigate the impact of target resolution on the reconstructed image, thereby exploring the effects of varying downsampling factors on the original image. We hypothesize that a higher downsampling factor will result in the loss of firstly structure-related information from the original image and therefore stronger style dependency of the feature map utilized for reconstruction. However, we also acknowledge the possibility of a threshold beyond which excessively high downsampling factors may lead to the loss of crucial style-related information. In this work we check the reconstruction resolutions 96, 48 and 24.

\textbf{Multi-Source Domain Generalization.} In addition to the experiments carried out in the SDG setting we also test our method against the baselines in the MSDG setting. The training is performed in the leave-one-domain-out manner where each domain is iteratively set out as the target domain and the rest serve as the source domains. Due to the fact that all of the methods involved are originally designed for SDG, no domain labels are utilized during the training and all of the source domains are effectively merged into one. We conduct MSDG experiments on Camelyon17-WILDS and DR datasets.

\textbf{Style augmentation.} We also study the impact of combining our method with various feature-map style-augmentation techniques. In the scope of this work we evaluate such methods as MixStyle \cite{zhou2021domain}, Domain Shift with Uncertainty (DSU) \cite{li2022uncertainty} and Correlated Style Uncertainty (CSU) \cite{zhang2024domain}.

\section{Results and Discussion}

\begin{table}
  \centering
  \scalebox{0.75}{
  \begin{tabular}{@{}clccccc@{}}
    \toprule
    Src & Method & $C_0$ & $C_1$ & $C_2$ & $C_3$ & $C_4$ \\
    \midrule
     & ERM & $-$ & $91.7_{\pm 0.1}$ & $89.8_{\pm 1.0}$ & $\mathbf{92.3_{\pm 0.9}}$ & \underline{$77.0_{\pm 2.0}$} \\
    
    $C_0$ & $\text{C}^2$SDG & $-$ & \underline{$91.8_{\pm 0.1}$} & \underline{$90.0_{\pm 0.8}$} & $92.1_{\pm 2.1}$ & $76.6_{\pm 2.1}$ \\
    
    & Ours & $-$ & $\mathbf{92.1_{\pm 0.1}}$ & $\mathbf{90.8_{\pm 0.6}}$ & \underline{$92.2_{\pm 0.6}$} & $\mathbf{80.9_{\pm 0.7}}$\\
    \midrule
    
    & ERM & \underline{$86.7_{\pm 1.7}$} & $-$ & \underline{$83.3_{\pm 1.4}$} & $\mathbf{88.3_{\pm 0.8}}$ & $71.0_{\pm 5.3}$ \\
    
    $C_1$ &$\text{C}^2$SDG & $85.4_{\pm 2.7}$ & $-$ & $82.9_{\pm 2.2}$ & $87.1_{\pm 4.0}$ & \underline{$71.8_{\pm 7.2}$} \\
    
    & Ours & $\mathbf{87.2_{\pm 0.3}}$ & $-$ & $\mathbf{84.3_{\pm 0.5}}$ & \underline{$88.1_{\pm 0.8}$} & $\mathbf{81.6_{\pm 1.3}}$\\
    \midrule
    & ERM 
    & \underline{$94.2_{\pm 0.1}$} 
    & \underline{$81.1_{\pm 1.0}$} 
    & $-$
    & \underline{$87.3_{\pm 0.3}$} 
    & \underline{$93.4_{\pm 0.1}$} \\
    
    $C_2$ & $\text{C}^2$SDG 
    & $93.9_{\pm 0.3}$ 
    & $80.5_{\pm 0.5}$ 
    & $-$
    & $86.6_{\pm 1.3}$ 
    & $91.8_{\pm 1.4}$ \\
    & Ours 
    & $\mathbf{94.3_{\pm 0.2}}$ 
    & $\mathbf{84.3_{\pm 0.3}}$ 
    & $-$
    & $\mathbf{88.5_{\pm 0.4}}$ 
    & $\mathbf{94.5_{\pm 0.6}}$\\
    \midrule
    & ERM 
    & \underline{$87.4_{\pm 0.8}$} 
    & \underline{$75.9_{\pm 0.6}$} 
    & \underline{$88.9_{\pm 0.7}$} 
    & $-$
    & \underline{$62.5_{\pm 3.6}$} \\
    $C_3$ & $\text{C}^2$SDG 
    & $85.3_{\pm 1.4}$ 
    & $75.2_{\pm 0.8}$ 
    & $88.4_{\pm 0.6}$ 
    & $-$
    & $59.8_{\pm 3.9}$ \\
    & Ours 
    & $\mathbf{88.6_{\pm 0.4}}$ 
    & $\mathbf{79.1_{\pm 0.8}}$ 
    & $\mathbf{89.9_{\pm 0.5}}$ 
    & $-$
    & $\mathbf{71.2_{\pm 0.7}}$\\
    \midrule
    & ERM 
    & $90.0_{\pm 0.6}$ 
    & $84.4_{\pm 0.7}$ 
    & $87.8_{\pm 0.6}$ 
    & \underline{$91.4_{\pm 0.6}$}
    & $-$\\
    $C_4$ & $\text{C}^2$SDG 
    & \underline{$90.7_{\pm 1.2}$} 
    & \underline{$84.9_{\pm 0.9}$} 
    & \underline{$88.3_{\pm 1.1}$} 
    & $89.9_{\pm 2.6}$
    & $-$\\
    & Ours 
    & $\mathbf{91.5_{\pm 0.5}}$ 
    & $\mathbf{86.6_{\pm 0.6}}$ 
    & $\mathbf{89.4_{\pm 0.3}}$ 
    & $\mathbf{92.2_{\pm 0.5}}$
    & $-$ \\
    \bottomrule
  \end{tabular}}
  \caption{Performance comparison on Camelyon17-WILDS in SDG setting. Each medical center number \(i\) is denoted as \(C_i\). The column Src represents the source domain and the columns \(C_0 - C_4\) - respective target domains. Each value shows accuracy  (\%) averaged over three runs with the standard deviation in the subscript}
  \label{tab:sdg_res_camelyon}
\end{table}

\begin{table}
  \centering
  \scalebox{0.8}{
  \begin{tabular}{@{}clcccc@{}}
    \toprule
    Src & Method & Aptos & EyePACS & Messidor & Messidor2  \\
    \midrule
     & ERM 
    & $-$
    & \underline{$65.2_{\pm 0.5}$} 
    & \underline{$46.3_{\pm 0.2}$} 
    & $58.2_{\pm 0.3}$ \\
    
    Aptos & $\text{C}^2$SDG 
    & $-$
    & $63.3_{\pm 1.8}$ 
    & $46.1_{\pm 0.2}$
    & \underline{$58.4_{\pm 0.2}$}  \\
    
    & Ours 
    & $-$
    & $\mathbf{65.7_{\pm 0.6}}$ 
    & $\mathbf{47.9_{\pm 0.3}}$ 
    & $\mathbf{59.3_{\pm 0.4}}$ \\
    \midrule
    
    & ERM 
    & $60.2_{\pm 0.4}$ 
    & $-$
    & \underline{$49.6_{\pm 0.3}$} 
    & \underline{$60.3_{\pm 0.4}$}  \\
    
    EyePACS & $\text{C}^2$SDG 
    & \underline{$60.6_{\pm 0.7}$} 
    & $-$
    & $49.3_{\pm 0.7}$
    & $60.0_{\pm 0.4}$  \\
    
    & Ours 
    & $\mathbf{61.9_{\pm 0.4}}$ 
    & $-$
    & $\mathbf{51.2_{\pm 0.7}}$ 
    & $\mathbf{60.9_{\pm 0.3}}$ \\
    \midrule
    & ERM 
    & \underline{$64.0_{\pm 0.7}$} 
    & \underline{$42.0_{\pm 2.8}$}
    & $-$
    & \underline{$59.6_{\pm 1.5}$}  \\
    
    Messidor &$\text{C}^2$SDG 
    & $62.4_{\pm 1.8}$ 
    & $39.7_{\pm 4.2}$
    & $-$
    & $59.3_{\pm 0.4}$  \\
    
    & Ours 
    & $\mathbf{64.0_{\pm 0.4}}$ 
    & $\mathbf{44.6_{\pm 0.3}}$ 
    & $-$
    & $\mathbf{59.8_{\pm 1.5}}$ \\
    \midrule
    & ERM 
    & $62.2_{\pm 2.0}$ 
    & $38.9_{\pm 2.5}$ 
    & $62.1_{\pm 2.0}$
    & $-$ \\
    
    Messidor2 & $\text{C}^2$SDG 
    & \underline{$64.2_{\pm 1.7}$} 
    & \underline{$41.8_{\pm 3.4}$}
    & \underline{$64.0_{\pm 1.1}$}
    & $-$ \\
    
    & Ours 
    & $\mathbf{67.7_{\pm 1.0}}$ 
    & $\mathbf{46.6_{\pm 1.3}}$ 
    & $\mathbf{65.0_{\pm 1.0}}$
    & $-$\\
    \bottomrule
  \end{tabular}}
  \caption{Performance comparison on DR datasets in SDG setting with the column Src representing source domain and the columns from 3rd to 6th representing target domains. Each value shows accuracy  (\%) averaged over three runs with the standard deviation in the subscript}
  \label{tab:sdg_res_dr}
\end{table}

\begin{table}[!t]
    \centering
    
    \setlength{\tabcolsep}{5.5pt}
    \renewcommand{\arraystretch}{1.1}
    
    \scalebox{0.7}{\begin{tabular}{lcccccc}\toprule[1.5pt]
         & $C_0$ & $C_1$ & $C_2$ & $C_3$ & $C_4$ & Average \\\midrule[1.25pt]
         
         
         ERM           
         & \underline{$96.5_{\pm 0.2}$}          
         & $89.1_{\pm 1.2}$            
         & \underline{$94.8_{\pm 0.5}$}      
         & \underline{$92.8_{\pm 0.4}$}   
         & $78.7_{\pm 2.3}$           
         & \underline{$90.4$}
         \\\midrule
         
         $\text{C}^2 \text{SDG}    $     
         & $96.3_{\pm 0.5}$          
         & \underline{$89.4_{\pm 2.0}$ }           
         & $94.3_{\pm 1.0}$      
         & $90.3_{\pm 2.6}$    
         & \underline{$79.7_{\pm 2.6}$}           
         & $90.0$
         \\\midrule
         
         Ours           
         & $\mathbf{97.3_{\pm 0.5}}$          
         & $\mathbf{92.1_{\pm 0.7}}$            
         & $\mathbf{96.1_{\pm 0.3}}$      
         & $\mathbf{94.2_{\pm 0.3}}$    
         & $\mathbf{84.1_{\pm 0.3}}$           
         & $\mathbf{92.8}$
         \\\bottomrule[1.5pt]
    \end{tabular}}
    \caption{Performance comparison in MSDG setting on Camelyon17-WILDS.  Each column represents accuracy on domain $C_i$ after training on the other 4 domains}
    \label{tab:msdg_table}
\end{table}

\begin{table}[!t]
    \centering
    \setlength{\tabcolsep}{5.5pt}
    \renewcommand{\arraystretch}{1.1}
    \scalebox{0.8}{\begin{tabular}{lccccc}\toprule[1.5pt]
         & Aptos & EyePACS & Messidor  & Messidor2 & Average \\\midrule[1.25pt]
         
         
         ERM           
         & \underline{$64.4_{\pm 1.4}$}          
         & \underline{$50.1_{\pm 0.9}$}          
         & \underline{$62.7_{\pm 1.8}$}      
         & \underline{$62.2_{\pm 0.5}$}           
         & \underline{$59.9$}
         \\\midrule
         
         $\text{C}^2 \text{SDG}    $     
         & $60.2_{\pm 2.3}$          
         &  $48.5_{\pm 1.8}$      
         & $61.7_{\pm 2.5}$    
         & $60.3_{\pm 1.6}$           
         & $57.7$
         \\\midrule
         
         Ours           
         & $\mathbf{66.1_{\pm 1.0}}$          
         & $\mathbf{51.8_{\pm 0.4}}$            
         & $\mathbf{63.0_{\pm 0.8}}$      
         &  $\mathbf{63.3_{\pm 0.8}}$           
         & $\mathbf{61.1}$
         \\\bottomrule[1.5pt]
    \end{tabular}}
    \caption{Performance comparison in MSDG setting on the composition of DR datasets.  Each column represents accuracy on the respective domain after training on the other 3 domains}
    \label{table:msdg_dr}

\end{table}

\begin{table*}[!t]
    \centering
    \caption{Evaluation of the effect of changing the reconstruction resolution in \ours{} is SDG setting. Each column represents performance after training on the domain $C_i$ averaged across the remaining 4 domains.}
    \label{table:res_table}
    \setlength{\tabcolsep}{6.5pt}
    \renewcommand{\arraystretch}{1.1}
    \begin{tabular}{lcccccc}\toprule[1.5pt]
         & $C_0$ & $C_1$ & $C_2$ & $C_3$ & $C_4$ & Average \\\midrule[1.25pt]
         
         96 \( \times \) 96           & $86.8_{\pm 0.9}$          & $\mathbf{83.0_{\pm 0.8}}$            & $88.9_{\pm 1.1}$      & $73.4_{\pm 0.4}$    & $88.6_{\pm 0.5}$           & $84.1$
         \\\midrule
         
         48 \( \times \) 48           & $\mathbf{87.4_{\pm 0.6}}$          & $81.8_{\pm 1.0}$            & $\mathbf{90.4_{\pm 0.7}}$      & $\mathbf{75.0_{\pm 0.8}}$    & $\mathbf{89.7_{\pm 0.6}}$           & $\mathbf{84.9}$
         \\\midrule
         
         24 \( \times \) 24           & $83.4_{\pm 0.7}$          & $81.3_{\pm 0.3}$            & $89.1_{\pm 0.3}$      & $73.1_{\pm 0.3}$    & $87.5_{\pm 1.5}$           & $82.9$
         
         \\\bottomrule[1.5pt]
    \end{tabular}

\end{table*}

\begin{table*}[!t]
    \centering
    \caption{Investigation of the effect of combining various style augmentation techniques with \ours{}. Each column represents performance after training on the domain $C_i$ averaged across the remaining 4 domains.}
    \label{table:style_table}
    \setlength{\tabcolsep}{5.5pt}
    \renewcommand{\arraystretch}{1.1}
    \begin{tabular}{lcccccc}\toprule[1.5pt]
         & $C_0$ & $C_1$ & $C_2$ & $C_3$ & $C_4$ & Average \\\midrule[1.25pt]
         
         MixStyle           & $85.2_{\pm 0.4}$          & $80.9_{\pm 0.8}$            & $\mathbf{92.1_{\pm 0.1}}$      & $73.7_{\pm 0.5}$    & $\mathbf{91.1_{\pm 0.3}}$           & $84.6$
         \\\midrule

         + Ours           & $\mathbf{87.2_{\pm 0.1}}$          & $\mathbf{82.8_{\pm 0.5}}$            & $\mathbf{92.1_{\pm 0.6}}$      & $\mathbf{75.6_{\pm 0.6}}$    & $90.6_{\pm 0.5}$           & $\mathbf{85.7}$
         \\\midrule[1.25pt]
         
         DSU           & $\mathbf{85.0_{\pm 0.7}}$          & $81.4_{\pm 0.8}$            & $\mathbf{91.1_{\pm 0.2}}$      & $\mathbf{76.4_{\pm 1.0}}$    & $\mathbf{90.2_{\pm 0.7}}$           & $84.8$

        \\\midrule

         + Ours           & $84.5_{\pm 0.7}$          & $\mathbf{84.2_{\pm 0.7}}$            & $90.7_{\pm 0.4}$      & $75.8_{\pm 0.8}$    & $89.5_{\pm 0.6}$           & $\mathbf{84.9}$
         
         \\\midrule[1.25pt]
         
         CSU           & $\mathbf{86.0_{\pm 0.5}}$          & $79.6_{\pm 2.1}$            & $90.7_{\pm 0.2}$      & $76.4_{\pm 0.6}$    & $\mathbf{90.4_{\pm 0.4}}$           & $84.6$
         
         \\\midrule

         + Ours           & $\mathbf{86.0_{\pm 0.4}}$          & $\mathbf{85.7_{\pm 0.5}}$            & $\mathbf{90.8_{\pm 0.7}}$      & $\mathbf{78.1_{\pm 0.6}}$    & $87.7_{\pm 0.9}$           & $\mathbf{85.7}$
         
         \\\bottomrule[1.5pt]
    \end{tabular}

\end{table*}

Table \ref{tab:sdg_res_camelyon} compares performance of our method versus ERM and $\text{C}^2$SDG \cite{hu2023devil} in SDG setting for each separate medical center being set as the source domain. Each value is achieved by averaging three experimental results with different seeds.

As it can be seen from the tables, our method consistently outperforms baselines by 1-2\% on average. Moreover it shows more stable performance with significantly smaller deviation across the runs. We argue that reconstruction-based regularization aids the disentanglement module of the framework by enforcing stronger dependency between the input images and style-dependent channels of the shallow-level feature map. Due to direct interconnection this improves extraction of the structure-related channels as well.

\begin{figure}[t]
\centering
    \includegraphics[width=0.45\textwidth]{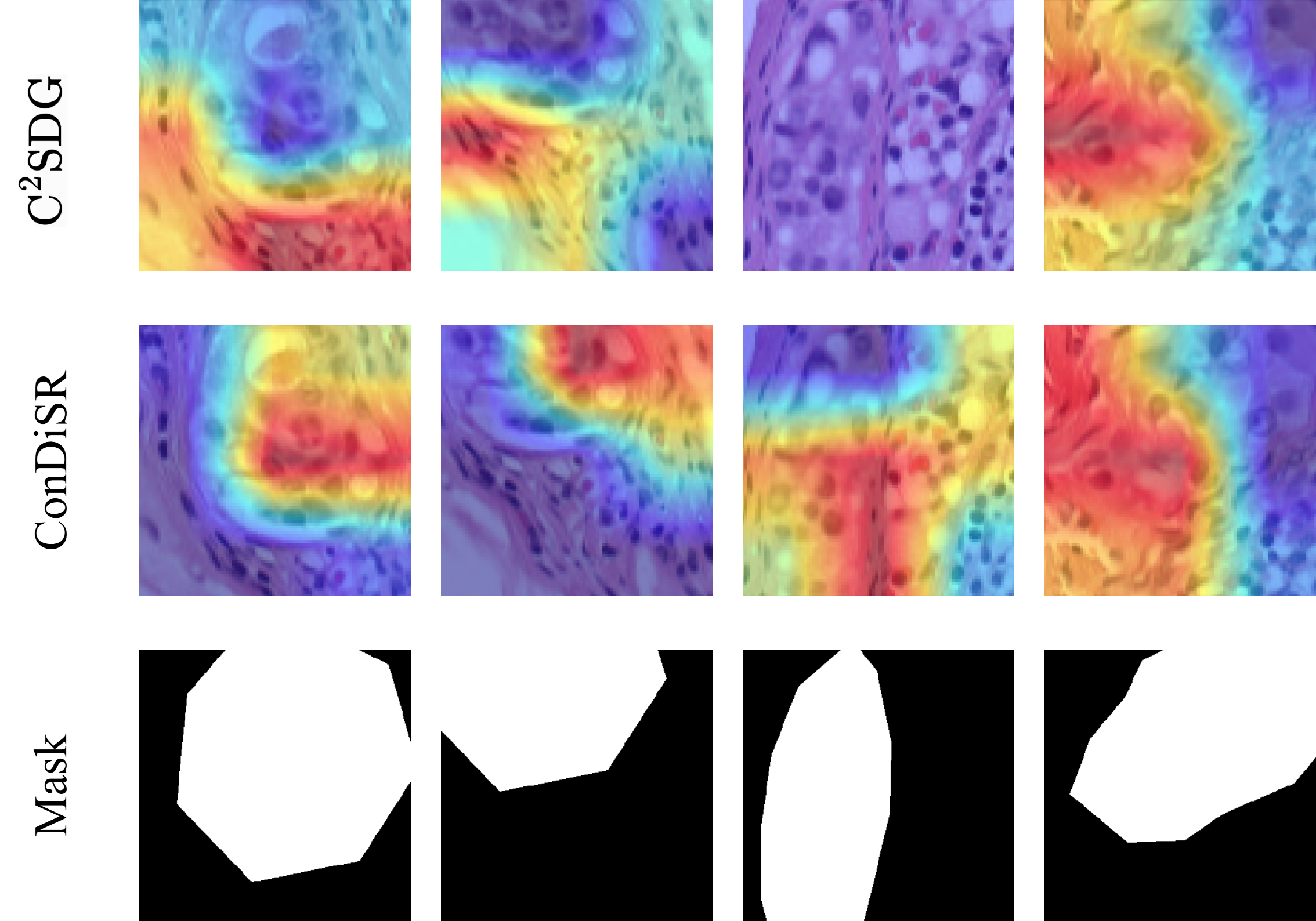} 
\caption{Qualitative performance comparison between \ours{} and C$^2$SDG \cite{hu2023devil} via Grad-CAM \cite{Selvaraju_2019} with the tumor presense masks taken from the original Camelyon17 challenge dataset \cite{bandi2018detection}.}
\label{fig:gradcam}
\end{figure}

The comparison between ERM and $\text{C}^2 \text{SDG}$ shows that both mehtods achieve very close performance scores, which leads us to believe that when it comes to the binary classification task, the primary driver of good performance of $\text{C}^2 \text{SDG}$ might be extensive augmentations as well as increased number of training iterations resulting from the augmentation strategy in \cite{hu2023devil} rather than the idea of channel-wise contrastive disentanglement in its default version.

The qualitative comparison of performance between $\text{C}^2 \text{SDG}$ and \ours{} is shown on the Figure \ref{fig:gradcam}. It is obtained by visualizing gradient activation maps \cite{Selvaraju_2019} on samples with positive tumor label. Comparison with segmentation masks from original Camelyon17 challenge dataset shows that our method pays more attention to areas, containing semantically important information.

Table \ref{tab:sdg_res_dr} compares performance of our method versus ERM and $\text{C}^2$SDG in SDG setting on the combination of the Diabetic Retinopathy datasets. Each of the datasets is one-by-one set as the source domain. As it can be seen from the tables, our method consistently outperforms the baselines while showing higher stability with lower deviation in results. The improvement is especially noticeable with Messidor2 as the source domain.

\textbf{Multi-Source Domain Generlaization.} The results of the experiments in MSDG setting are shown in the Tables \ref{tab:msdg_table} and \ref{table:msdg_dr}. As it can be seen from the Table \ref{tab:msdg_table} when applied to Camelyon17-WILDS our method noticeably outperforms baselines on each of the tagret domains. The difference is epecially noticeable on the data from the medical center 4, which overall is the hardest to generalize to for all of the methods. The average accuracy improvement is from \(90.4\%\) to \(92.8\%\) which corresponds to approximately \(20\%\) classification error reduction. When it comes to DR grading, our method again outperforms the baselines by the margin of more than \(1\%\).

\textbf{Reconstruction resolution.} From the Table \ref{table:res_table} we can see that the optimal reconstruction resolution for our method is 48x48. This supports our preliminary assumption, that reconstruction of the input images of the original size (96x96) forces unnecessary structural features into \( f_{sty} \) and hampers the work of disentanglement module. On the other hand, reconstruction of the images resized to the resolutions that are too small (24x24) does not help the model's performance. Similarly to \cite{he2022masked}, the gap seems to arise between the tasks of recognition and reconstruction. When the target image resolution is too low, the decoder of the reconstruction model is too shallow (as it consists of fewer upsampling blocks) which makes the input of the reconstruction network, \(f_{sty}\), too specialized for reconstruction but not particularly informative from the point of view of recognition. 

It is important to clarify, that Table \ref{table:res_table} shows performance in the SDG setting with each column representing the source domain and the accuracy being computed over the rest of the domains unlike it is presented in the Tables \ref{tab:sdg_res_camelyon} and \ref{tab:sdg_res_dr}.

\textbf{Style augmentation.} As it can be seen in the Table \ref{table:style_table}, application of various style-augmentation methods \cite{zhou2021domain} to the feature maps after each of the first three layers of the network does not provide any significant improvement compared to the baselines. However in conjunction with \ours{}, it results in the highest accuracy reached, which raises an assumption that despite the usage of the disentanglement technique, structure-related feature maps extracted by our method still carry various important style-dependent features that are correlated with the classification label and benefit from augmentations. Similarly to Table \ref{table:res_table}, Table \ref{table:style_table} shows performance in the SDG setting with each column representing the source domain and the accuracy being computed over the rest of the domains.

\section{Conclusion}
This paper presents a novel method, \ours{}, designed for single domain generalization in the classification of images. Leveraging contrastive channel-wise disentanglement alongside style regularization through low-resolution reconstruction, \ours{} demonstrates an improvement over baseline methods, achieving an increase in accuracy in both SDG and MSDG settings on various medical imaging modalitites. Successful application of \ours{} shows the potential of disentanglement-based methodologies in enhancing model performance and generalizability, particularly in classification tasks within complex and heterogeneous datasets such as histopathology images. Our results underscore the importance of further exploration of disentanglement-based approaches and encourage further research in this domain. Potential future work may involve evaluation of the proposed method's performance on other classification datasets in order to test its generalizability.

{\small
\bibliographystyle{ieee_fullname}
\bibliography{egbib}
}

\end{document}